\documentclass[final]{cvpr}

\usepackage{times}
\usepackage{epsfig}
\usepackage{graphicx}
\usepackage{amsmath}
\usepackage{amssymb}
\usepackage{booktabs}
\usepackage{enumitem}
\usepackage{physics}
\usepackage{multirow}

\makeatletter
\@namedef{ver@everyshi.sty}{}
\makeatother
\usepackage{tikz}
\usepackage{pgfplots}
\usepackage{tabularray}
\usepackage{hhline}
\usepackage[page]{appendix}
\usepackage{subcaption}
\usepackage{makecell} 
\usepackage{graphicx}
\usepackage{framed}

\usepackage{float}
\usepackage{listings}
\usepackage[frozencache,cachedir=./mint/]{minted}

\usepackage[pagebackref=true,breaklinks=true,colorlinks,bookmarks=false]{hyperref}

\begin{document}

\title{Code Generation with AlphaCodium: From Prompt Engineering to Flow Engineering}

\author{Tal Ridnik, Dedy Kredo, Itamar Friedman\\
CodiumAI\\
{\tt\small \{tal.r, dedy.k, itamar.f\}@codium.ai}
}

\maketitle

\begin{abstract}
\label{sec:abstract}
Code generation problems differ from common natural language problems - they require matching the exact syntax of the target language, identifying happy paths and edge cases, paying attention to numerous small details in the problem spec, and addressing other code-specific issues and requirements. Hence, many of the optimizations and tricks that have been successful in natural language generation may not be effective for code tasks.
In this work, we propose a new approach to code generation by LLMs, which we call AlphaCodium - a test-based, multi-stage, code-oriented iterative flow, that improves the performances of LLMs on code problems.
We tested AlphaCodium on a challenging code generation dataset called CodeContests, which includes competitive programming problems from platforms such as Codeforces. The proposed flow consistently and significantly improves results.
On the validation set, for example, GPT-4 accuracy (pass@5) increased from 19\% with a single well-designed direct prompt to 44\% with the AlphaCodium flow. 
Many of the principles and best practices acquired in this work, we believe, are broadly applicable to general code generation tasks.

Full implementation is available at: \url{https://github.com/Codium-ai/AlphaCodium}

\end{abstract}

\section{Introduction}
\begin{figure*}[t!]
\centering
\begin{subfigure}[a]{.99\textwidth }
  \centering
  \includegraphics[scale=0.60]{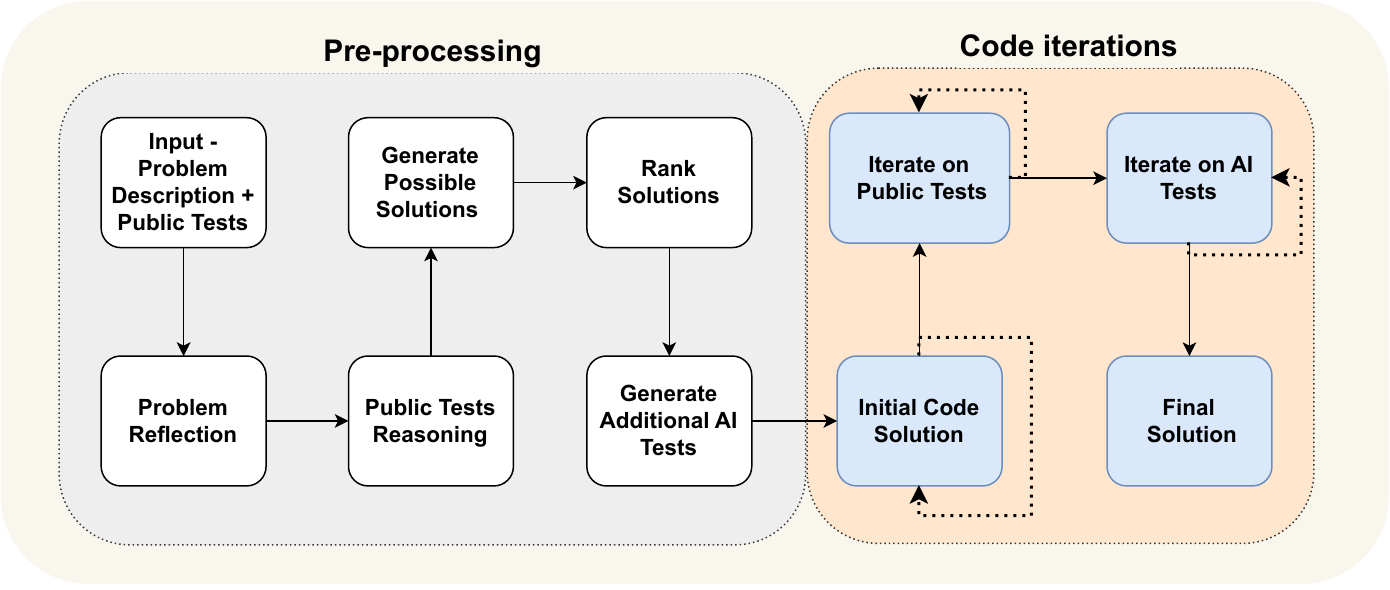}
  \caption{The proposed AlphaCodium flow.}
  \vspace{5mm}%
\end{subfigure}
\begin{subfigure}[a]{.99\textwidth }
  \centering
  \includegraphics[scale=0.70]{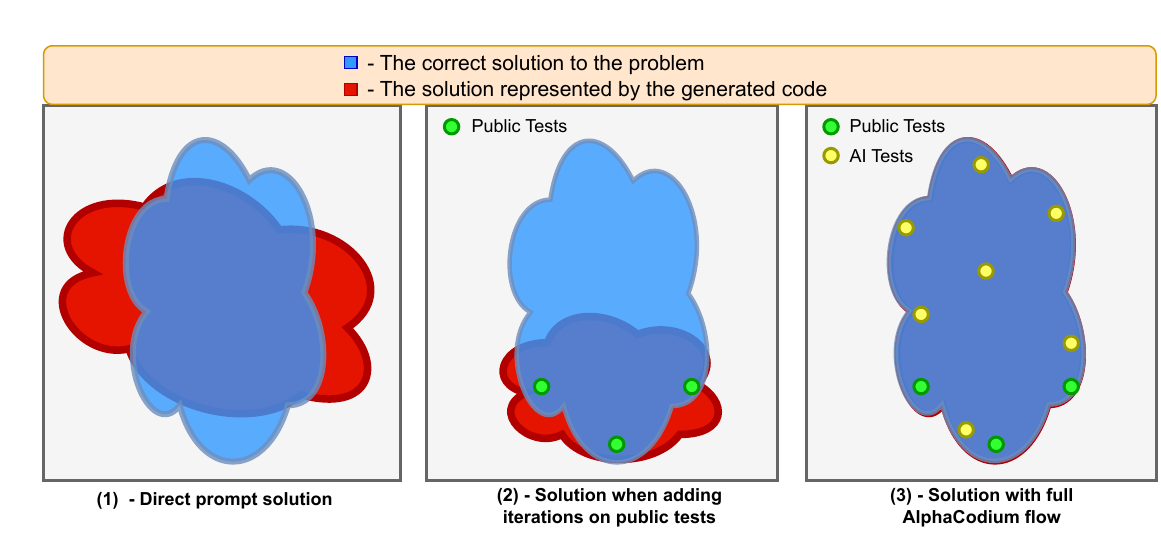}
  \caption{Illustrating the improvement from AlphaCodium.}
\end{subfigure}

\caption{\textbf{Illustration of AlphaCodium flow contribution} - with direct prompt, the model struggles to solve code problems. Iterating on public tests stabilizes and improves the solution but leaves "blind spots" because the public tests are not comprehensive. The full AlphaCodium flow, which includes a pre-processing phase as well as iterations on public and AI-generated tests, allows the solution to be further improved, leading to increased solve ratio.}
\label{fig:alpha_codium_flow}
\end{figure*}

With a sparse reward signal, code generation tasks require searching in the huge structured space of possible
programs. Correct solutions to the same problem can look significantly different, and judging if a partial or incorrect solution is useful is a difficult challenge - a single-character edit can completely alter the solution's behavior.  Due to the unique nature of code generation tasks, common prompting techniques that have been optimized for natural language tasks~\cite{dhuliawala2023chain,wang2022self, nori2023can}, may not be as effective when applied to code generation.

Recent large-scale transformer-based language models~\cite{vaswani2017attention} have successfully generated code that solves simple programming tasks~\cite{chen2021evaluating,austin2021program}. However, real-world code problems are often different in nature - they are more nuanced, and can be defined by a long natural language task description (i.e., spec), that contains multiple details and rules that the solution code must address.

The introduction of CodeContests~\cite{li2022competition}, a dataset curated from competitive programming platforms such as Codeforces~\cite{mirzayanov2020codeforces}, enabled the evaluation of models and flows on more challenging code problems, which usually include a lengthy problem description. A private test set, with more than 200 unseen tests per problem, enables to evaluate the generated code comprehensively, and to reduce false positive rates to a minimum.

The primary work addressing the CodeContests dataset was AlphaCode~\cite{li2022competition}, a code generation system developed by DeepMind, that utilizes a fine-tuned network specifically for competitive programming tasks. AlphaCode generates a very large number of possible solutions (up to 1M), that are then processed and clustered, and among them a small number ($\sim$ 10) is chosen and submitted. While the results of AlphaCode are impressive, the need to fine-tune a model specifically for code-oriented tasks, and the heavy computational brute-force-like load, makes it impractical for most real-life usages.
CodeChain~\cite{le2023codechain} is another work to tackle competitive programming tasks, which introduced a novel inference framework to improve code generation in LLMs through a chain of sub-module-based self-revisions.

In this paper, we present AlphaCodium, a code-oriented flow that revolves around an iterative process where we repeatedly run and fix a generated code against input-output tests. Two key elements for AlphaCodium flow are (a) generating additional data, such as problem reflection and test reasoning, to aid the iterative process, and (b) enrichment of public tests with additional AI-generated tests. The proposed flow, which is depicted in Figure~\ref{fig:alpha_codium_flow}, is divided into two main phases: a pre-processing phase where we reason about the problem in natural language, and an iterative code generation phase where we generate, run, and fix a code solution against public and AI-generated tests.

A key observation when designing AlphaCodium flow is that generating additional useful tests is easier than generating a correct code solution. Adding specific tests requires mainly understanding the problem, some insight, and basic brute-force or logical reasoning. There is no need to fully “solve” the problem when generating additional tests.

AlphaCodium flow also utilizes novel code-oriented design concepts, tricks, and best practices, such as:
(1) YAML structured output; (2) bullet point analysis to encourage semantic reasoning; (3) generating modular code; (4) soft decisions with double validation; (5) encouraging exploration and postponing direct decisions; (6) test anchors.

AlphaCodium's flow, when compared against a well-designed single prompt,  consistently and significantly improves the performance of LLMs on CodeContests problems. This is true both for open-source models
(DeepSeek~\cite{deepseek-coder}) and closed-source models (GPT~\cite{floridi2020gpt}). For GPT-4 on the validation set,
for example, the pass@5 accuracy improved from 19\% to 44\%. AlphaCodium also outperforms previous works, while having a significantly smaller computational budget - it achieves superior results over AlphaCode, for example, with four orders of magnitude fewer LLM calls.

We believe many of the principles and best practices used in this work broadly apply to general code generation tasks. In addition, we argue that utilizing harder and more complicated benchmarks like CodeContests dataset will allow the community to better evaluate LLMs, compared to more simplistic benchmarks, which are common today, like HumanEval~\cite{chen2021evaluating}.

\section{CodeContests Dataset}
\begin{figure*}[t!]
\centering
\begin{subfigure}[a]{.99\textwidth }
  \centering
  \includegraphics[scale=0.85]{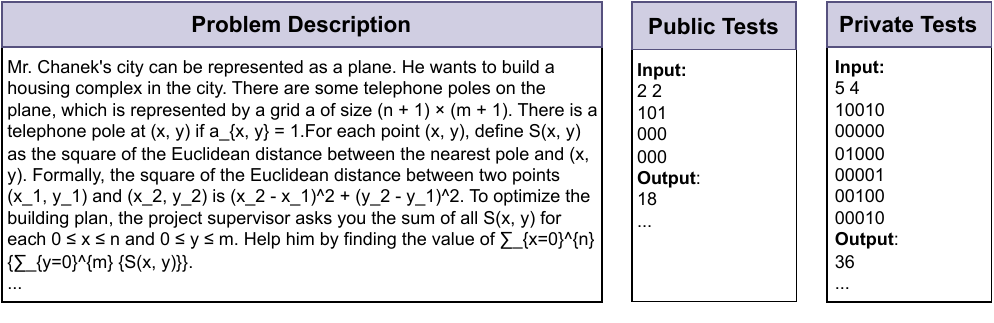}
  \caption{A typical CodeContests problem.}
  \vspace{5mm}%
\end{subfigure}
\begin{subfigure}[a]{.99\textwidth }
  \centering
  \includegraphics[scale=0.80]{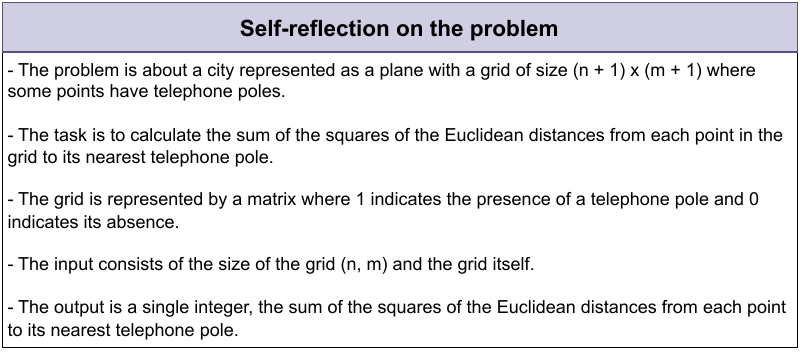}
  \caption{An AI-generated self-reflection on the problem.}
\end{subfigure}
\caption{\textbf{Problem description and reflection} - an example of a typical CodeContests problem, with AI-generated self-reflection on the problem. While the original description is lengthy and complicated, proper self-reflection makes the problem clearer and more coherent, leading to improved code solutions.}
\label{fig:typical_problem}
\end{figure*}

CodeContests~\cite{li2022competition} is a challenging code generation dataset introduced by Google’s DeepMind, involving problems curated from competitive programming platforms such as Codeforces~\cite{mirzayanov2020codeforces}.
The dataset contains ~10K code problems that can be used to train LLMs, as well as a validation and test set to assess the ability of LLMs to solve challenging code problems.

In this work, instead of training a dedicated model, we focused on developing a code-oriented flow, that can be applied to any LLM pre-trained to support coding tasks, such as GPT~\cite{floridi2020gpt} or DeepSeek~\cite{deepseek-coder}. Hence, we chose to ignore the train set, and focused on the validation and test sets of CodeContests, which contain 107 and 165 problems, respectively. Figure \ref{fig:typical_problem}(a) depicts an example of a typical problem from CodeContests dataset.
Each problem consists of a description and public tests, available as inputs to the model. The goal is to generate a code solution that produces the correct output for any (legal) input. A private test set, which is not available to the model or contesters, is used to evaluate the submitted code solutions.

\paragraph{What makes CodeContests a good dataset for evaluating LLMs on code generation tasks:}
\hfill

1) CodeContests, unlike many other competitive programming datasets~\cite{hendrycks2021measuring, chen2021evaluating}, utilizes a comprehensive private set of tests to avoid false positives - each problem contains $\sim$200 private input-output tests the generated code solution must pass.

2) LLMs generally do not excel at paying attention to small details, because they typically transform the problem description to some "average" description, similar to common cases on which they were trained. Real-world code problems, on the other hand, frequently contain minor details that are critical to their proper resolution. A key feature of the CodeContests dataset is that the problem descriptions are, by design, complicated and lengthy, with small details and nuances (see a typical description in Figure \ref{fig:typical_problem}(a)). We feel that adding this degree of freedom of problem understanding is beneficial since it simulates real-life problems, which are often complicated and involve multiple factors and considerations.  This is in contrast to more common code datasets such as HumanEval~\cite{chen2021evaluating}, where the problems are easier, and presented in a concise manner. An example of a typical HumanEval problem appears in appendix \ref{app:human_eval}.

Figure \ref{fig:typical_problem}(b) depicts the model's introspection on the problem presented in Figure \ref{fig:typical_problem}(a). Note that proper self-reflection makes the problem clearer and more coherent. This illustrates the importance of problem understanding as part of a flow that can lead, with high probability, to generating a correct code solution.

\section{The Proposed Flow}
\subsection{Overview}
Due to the complicated nature of code generation problems, we observed that single-prompt optimizations, or even chain-of-thought prompts, have not led to meaningful improvement in the accuracy of LLMs on CodeContests. The model struggles to understand and comprehend the problem, and continuously produces wrong code, or a code that passed public tests but fails to generalize to unseen private tests. Common flows, that are suitable for natural language tasks, may not be optimal for code-generation tasks, which include an untapped potential - repeatedly \textbf{running} the generated code, and validating it against known examples.

Instead of common prompt engineering techniques used in NLP, we found that to solve CodeContest problems it was more beneficial to employ a dedicated code-generation and testing-oriented flow, that revolves around an iterative process where we repeatedly run and fix the generated code against input-output tests (see Figure~\ref{fig:typical_problem}(a) for examples of such tests). Two key elements for this code-oriented flow are (a) generating additional data in a pre-processing stage, such as self-reflection and public tests reasoning, to aid the iterative process, and (b) enrichment of the public tests with additional AI-generated tests. 

In Figure \ref{fig:alpha_codium_flow}(a) we present AlphaCodium flow for solving competitive programming problems. The flow is divided into two main phases:
\begin{itemize}
    \item 
    The \textit{pre-processing} phase represents a linear flow where AlphaCodium reasons about the problem, in natural language,
    \item 
    The \textit{code iterations} phase includes iterative stages where AlphaCodium generates, runs, and fixes a solution code against certain tests. 

\end{itemize}

\subsection{Flow stages}
In this section, we will review the different stages used in AlphaCodium flow (Figure~\ref{fig:alpha_codium_flow}(a)):

\hrulefill
\vspace{-2mm}
\paragraph{Problem reflection} Describe the problem, in bullet points, while addressing the problem goal, inputs, outputs, rules, constraints, and other relevant details that appear in the problem description.

\hrulefill
\vspace{-2mm}
\paragraph{Public tests reasoning} Explain why each test input leads to the output.

\hrulefill
\vspace{-2mm}
\paragraph{Generate possible solutions} Generate a list of 2-3 possible solutions to the problem, described in natural language.

\hrulefill
\vspace{-2mm}
\paragraph{Rank solutions} Rank the possible solutions and choose the “best solution”, in terms of correctness, simplicity, and robustness. (not necessarily take the “most efficient” solution).

\hrulefill
\vspace{-2mm}
\paragraph{Generate additional AI tests} Generate an additional 6-8 diverse input-output tests for the problem. 
Try to cover cases and aspects not covered by the original public tests.

\hrulefill
\vspace{-2mm}
\paragraph{Initial code solution} The goal of this stage is to generate an initial code solution to the problem. It is essential that this code will reasonably "close" to the correct code, so the run-fix iterations in the next stages will have a better chance of succeeding.
The stage flow:
\begin{itemize}
    \item Choose a potential solution. Generate a corresponding code, and run it on selected public and AI tests.
\item Repeat this process until the tests pass, or until a try-limit is reached.
\item The first code that passes the tests, or the code with the closest output (see appendix~\ref{app:distance_between_tests}), will be used as the base code for the next steps.
\end{itemize}

\hrulefill
\vspace{-2mm}
\paragraph{Iterate on public tests} Start from the base code. Iteratively run it on the public tests. If the code fails on a specific test, try to fix it, given the error message.

\hrulefill
\vspace{-2mm}
\paragraph{Iterate on AI-generated Tests} Continue the run-fix iterations on the AI-generated tests. Use “test anchors” (see section~\ref{sec:design_concepts}).

\hrulefill
\\
\subsection{Additional insights}

In this section, we will offer additional insights and share intuitions about the proposed flow.

Firstly, the flow relies on knowledge accumulation - trying to progress from easy to hard, gaining knowledge and insight along the way to help with the more difficult stages. For example, the output of the first step, \textit{problem reflection}, can be utilized as prompt input to more difficult steps like \textit{generate possible solutions}. The pre-processing phase's outputs are used to aid the most challenging and critical phase, code iterations, where we try to generate code that correctly solves the problem.

Another key observation in designing AlphaCodium is that for AI, generating more tests is easier than generating a full solution code. Generating additional tests requires mainly understanding the problem and basic brute-force or logical reasoning. There is no need to fully “solve” the problem in order to generate additional useful input-output test pairs. This is in contrast to generating a correct solution code, which requires a complete algorithmic solution, equivalent to correctly solving any possible pair of input-output tests. As a result, we can generate more AI tests, and then leverage them to improve the code creation phase, as described in Figure \ref{fig:alpha_codium_flow}(b). We further amplify the contribution of these additional tests by asking the model to focus on aspects not addressed by the original public tests, such as large inputs, edge cases, and so on.

Also note that some steps can be combined into a single LLM call, and the flow in Figure \ref{fig:typical_problem}(a) is a conceptual flow, emphasizing the process's high-level steps. In practice, structured output (see section~\ref{sec:design_concepts}) enables to combine multiple stages into a single LLM call, in order to save resources, or because a model performs better when doing specific tasks concurrently.

\section{Code-Oriented Design Concepts}
\label{sec:design_concepts}
\begin{figure*}[hbt!]
\begin{framed}

\begin{minted}[fontsize=\small]{text}

...
Your goal is to present possible solutions to the problem.
Make sure that each solution fully addresses the problem goals, rules, and 
constraints.


The output must be a YAML object equivalent to type $PossibleSolutions, according to 
the following Pydantic definitions:
\end{minted}
\begin{minted}[fontsize=\small]{python}
class Solution(BaseModel):
   name: str = Field(description="The name of the solution")
   content: str = Field(description="A description of the solution")
   why_it_works: str = Field(description="Why this solution is correct. Be specific\ 
   and detailed regarding the problem rules and goals")
   complexity: str = Field(description="The complexity of the solution")

class PossibleSolutions(BaseModel):
   possible_solutions: List[Solution] = Field(max_items=3, description="A list of\
   possible solutions to the problem. Make sure each solution fully addresses the\
   problem rules and goals, and has a reasonable runtime - less than three seconds\
   on a modern computer, given the problem constraints for large inputs.")
\end{minted}
\end{framed}
\caption{\textbf{Example for a prompt with structured output} (\textit{generate possible solutions} stage)}
\label{fig:structerd_output}
\end{figure*}

In this section we will present additional design concepts, tricks, and best practices we found beneficial when trying to solve code generation problems. AlphaCodium flow proposed in Figure \ref{fig:alpha_codium_flow} extensively uses these design concepts.

\paragraph{YAML Structured output:} the usage of structured output - asking the model to generate an output in YAML format, equivalent to a given Pydantic class - is a key component in our proposed flow. An example of such instruction (\textit{possible solutions} stage) appears in Figure \ref{fig:structerd_output}.

Structured output eliminates the majority of the hassle and dark knowledge required for "prompt engineering" and instead allows complicated tasks to be presented in a straightforward, code-like manner. It also makes it possible to obtain complex answers that involve several stages,  representing a logical and methodical thinking process. 

While newer versions of GPT models~\cite{floridi2020gpt} have built-in support for JSON-style output, we argue that YAML output is far more suitable specifically for code generation tasks, see appendix~\ref{app:yaml_vs_json}.

\paragraph{Semantic reasoning via bullet points analysis:} when asking an LLM to reason about an issue, better results are obtained when demanding the output to be in bullet points format. Bullet points encourage an in-depth understanding of the problem, and force the model to divide the output into logical semantic sections, leading to improved results. For example, with self-reflection on a problem in bullet points (Figure~\ref{fig:typical_problem} (b)), each bullet point represents a semantic understanding of a different part of the problem - general description, goals and rules, input structure, and output structure. 

\paragraph{LLMs do better when generating a modular code:} when LLMs are asked to generate a single lengthy function, we observed poor results - the code often contains bugs or logical mistakes. Even worse, a single monolithic code hurts the ability to perform iterative fixing - the model struggles to pinpoint and fix problems, even when given the error message. 
When clearly asking the model to: “\textit{divide the generated code into small sub-functions, with meaningful names and functionality}”, we observe a better-produced code, with fewer bugs, and higher success rates for the iterative fixing stages.

\paragraph{Soft decisions with double validation:} LLMs tend to struggle with code tasks that require them to think, reason, and make strict, non-trivial decisions. Let’s take for example the task of generating additional tests for a problem. Quite often, some of the tests the model generates will be plain wrong. With a \textit{double validation} process, we add an extra step where, given the generated output, the model is asked to re-generate the same output, but correct it if needed. For example, given the generated AI tests as input, the model is asked to re-generate the same tests, while correcting wrong output, if exists. We found that this step of double validation, while encouraging the model to be critical and to reason, is more effective than asking a direct yes\textbackslash no question: "is this test correct?"

\paragraph{Postpone decisions, try to avoid direct questions, and leave room for exploration:} when we ask the model direct questions regarding complicated issues, we consistently see hallucinations and wrong answers. To counter this, we adopt a flow of gradual data accumulation, from easier tasks to harder ones:
\begin{itemize}
\item 
Start with the easiest tasks - self-reflection on the problem, and reasoning about public tests.
\item
Move to generating additional AI tests, and possible solutions to the problem
\item
Only after we acquire the model's answers to the tasks above, we move to actual code generation, and run-fix iterations.
\end{itemize}
As another example, instead of choosing a single algorithmic solution to the problem, we prefer to rank several possible solutions, and give priority, but not exclusiveness, to the top-ranked solution when generating initial code. Since the model can be wrong, it's better to avoid irreversible decisions, and leave room for exploration and code iterations with different possible solutions.

\paragraph{Test anchors:} even with double validation, some AI-generated tests will be wrong. This makes iterations on them challenging - when a test fails, how can we know if it is because the code is wrong, or because the test is wrong? When we ask the model directly “who is wrong”, we often see hallucinations, and may end up with a  wrongly fixed code. To tackle this problem, we utilized a technique of \textit{test anchors}:
\begin{itemize}
\item 
Iterate first on the public tests, which we know are correct. When finished, set all the passed tests as anchor tests.
\item 
Now iterate on the AI-generated tests, one by one. If a test passes, add it to the list of test anchors
\item 
If a test fails, assume it's because the code is incorrect, and try to fix the code. However, demand that the fixed code will also pass all the test anchors already acquired. As a result, the test anchors will protect us against an incorrectly fixed code.
\end{itemize}

Another optimization for test anchors is to sort the AI-generated tests from easy to hard. That way, there are more chances that the iterative process will acquire anchors at the beginning of the process, which can be used as protection later when iterating on the more complicated AI tests.

\paragraph{What did not work for us:} In appendix~\ref{app:what_didnt_work} we present additional tricks and methods we tried, which have not led to improved results.

\section{Results}
\label{sec:results}
\subsection{Direct prompt vs. AlphaCodium flow}
In Table~\ref{table:direct_prompt} we compare AlphaCodium results to the results obtained with a single well-designed direct prompt. The metric being used is pass@k, defined as the percentage of problems solved by
using k generated solutions per problem.
\begin{table}[hbt!]
\centering
\begin{tblr}{
  width = \linewidth,
  colspec = {Q[194]Q[177]Q[246]Q[183]},
  cell{2}{1} = {r=4}{},
  cell{2}{2} = {r=2}{},
  cell{4}{2} = {r=2}{},
  cell{6}{1} = {r=4}{},
  cell{6}{2} = {r=2}{},
  cell{8}{2} = {r=2}{},
  cell{10}{1} = {r=4}{},
  cell{10}{2} = {r=2}{},
  cell{12}{2} = {r=2}{},
  vline{2-4} = {1-24,6,10}{},
  vline{4} = {3,5,7,9,11,13}{},
  vline{3-4} = {4,8,12}{},
  hline{2,6,10} = {-}{},
  hline{4,8,12} = {2-4}{},
}
model    & set        & method        & {score \\(pass@5)} \\
{ DeepSeek\\\centering{-33B}~\cite{deepseek-coder}} & validation & Direct        & 7\%                \\
         &            & AlphaCodium   & \textbf{20\%}               \\
         & test       & Direct prompt & 12\%               \\
         &            & AlphaCodium   & \textbf{24\%}               \\
         \hline
GPT-3.5  & validation & Direct prompt & 15\%               \\
         &            & AlphaCodium   & \textbf{25\%}               \\
         & test       & Direct prompt & 8\%                \\
         &            & AlphaCodium   & \textbf{17\%}               \\
         \hline
GPT-4    & validation & Direct prompt & 19\%               \\
         &            & AlphaCodium   & \textbf{44\%}               \\
         & test       & Direct prompt & 12\%               \\
         &            & AlphaCodium   & \textbf{29\%}               
\end{tblr}
\caption{\textbf{Comparison of AlphaCodium flow results to direct prompt on various models.}}
\label{table:direct_prompt}
\end{table}
As can be seen, AlphaCodium flow consistently and significantly improves the performance of LLMs on CodeContests problems. This is true both for open-source (DeepSeek) and close-source (GPT) models, and for both the validation and test sets. For GPT-4 on the validation set, for example, the pass@5 score improves from 19\% to 44\% - x2.3 improvement.

\subsection{Comparison to previous works}
In Table~\ref{table:comparison_to_others} we compare AlphaCodium results to other methods from the literature.
\begin{table}
\centering
\begin{tblr}{
  width = \linewidth,
  colspec = {Q[227]Q[196]Q[365]Q[121]},
  cell{2}{1} = {r=4}{},
  cell{2}{2} = {r=2}{},
  cell{4}{2} = {r=2}{},
  cell{6}{2} = {r=3}{},
  cell{7}{1} = {r=2}{},
  cell{9}{2} = {r=3}{},
  cell{10}{1} = {r=2}{},
  vline{4} = {1-22}{},
  vline{3} = {1-22}{},
  vline{2} = {1-22}{},
  hline{2,6,9} = {-}{},
  hline{3,5} = {3-4}{},
  hline{4} = {2-4}{},
  hline{7,10} = {1,3-4}{},
}
model     & set        & method                      & score         \\
GPT-3.5   & validation & {AlphaCodium\\(pass@5)}     & \textbf{25\%} \\
          &            & {CodeChain\\(pass@5)}       & 17\%          \\
          & test       & {AlphaCodium\\(pass@5)}     & \textbf{17\%}          \\
          &            & {CodeChain\\(pass@5)}       & 14\%          \\
\hline
GPT-4     & validation & {AlphaCodium\\(pass@5)}     & \textbf{44\%} \\
AlphaCode &            & {AlphaCode\\(pass@10@1K)}   & 17\%          \\
          &            & {AlphaCode\\(pass@10@100K)} & 24\%          \\
\hline
GPT-4     & test       & {AlphaCodium\\(pass@5)}     & \textbf{29\%} \\
AlphaCode &            & {AlphaCode\\(pass@10@1K)}   & 16\%          \\
          &            & {AlphaCode\\(pass@10@100K)} & 28\%          
\end{tblr}
\caption{\textbf{Comparison of AlphaCodium to other works from the literature.}}
\label{table:comparison_to_others}
\end{table}
As can be seen, when comparing AlphaCodium to CodeChain with the same model (GPT-3.5) and the same metric (pass@5), AlphaCodium consistently does better.

When comparing AlphaCodium to AlphaCode work, we need to take into account that AlphaCode uses a different generation methodology - fine-tuning an (unknown) model specifically for code problems, generating a very large number of code solutions, clustering them, and submitting K solutions from the top clusters. pass@10@100K, for example, means the 100K (!) solutions were generated and clustered, and 10 solutions were finally chosen and submitted. AlphaCode used a fine-tuned model, and utilized a brute-force-like approach with a significantly higher number of LLM calls. Yet, the top results achieved by AlphaCodium are better

Note that neither AlphaCode nor CodeChain papers~\cite{li2022competition,le2023codechain} released a reproducible open-source solution for CodeContests, including end-to-end generation and evaluation scripts. There are subtleties when evaluating results. For example - how to treat problems with multiple solutions, how to address tolerance issues, timeouts, and more.
We compare to the numbers reported in the papers, but release a full reproducible code and evaluation script of AlphaCodium, to enable future comparisons to be more reliable and consistent.

\subsection{Computational effort and comparison to AlphaCode and AlphaCode2}
With AlphaCodium flow we perform $\sim$15-20 LLM calls per solution, so a pass@5 submission involves $\sim$ 100 LLM calls.

AlphaCode did not report how many LLM calls were done per run~\cite{li2022competition}. Let's assume one call per run was done (unknown, could be more), then a pass@10@100K (i.e. ten submissions, curated from 100,000 generated solutions) involves 1M LLM calls, four orders of magnitude more than AlphaCodium. Yet, the top results obtained by AlphaCodium are better.

Recently, a new work called AlphaCode2~\cite{team2023gemini} was published, where a Gemini-Pro model was fine-tuned and evaluated on code programming problems.  The paper also reported results on a CodeContests benchmark, but on an updated variant that they did not release to the public. 
According to AlphaCode2 report: “AlphaCode2 requires about 100 samples to reach the level of performance of AlphaCode with a million samples, making it over 10000× more sample efficient.”
Hence both AlphaCode2 and AlphaCodium are four orders of magnitude more efficient than AlphaCode, LLMs call-wise.
But, AlphaCode2 utilized a modern foundation model (Gemini-Pro) that was heavily fine-tuned specifically for CodeContests competition, while AlphaCodium uses general-purpose models as-is, and improves their performances without extra data and an expensive training phase.

\section{Conclusions}
In this paper, we introduced AlphaCodium, a code-oriented flow that iteratively runs and fixes a generated code against input-output tests.
The flow is divided into two main phases: a pre-processing phase, where AlphaCodium reasons about the problem in natural language, and a code iterations
phase, in which AlphaCodium iterates on public and AI-generated tests.

AlphaCodium also utilizes additional design concepts, tricks, and best practices we found beneficial for code generation: structured output in YAML format, generating modular code, semantic reasoning via bullet point analysis, soft decisions with double validation, encouraging exploration, and test anchors.

We tested AlphaCodium on a challenging code generation dataset called CodeContests. The proposed flow consistently and significantly improves results of various closed-source and open-source models. AlphaCodium also outperforms previous works from the literature, while having a significantly smaller computational budget.

{\small
\bibliographystyle{ieee_fullname.bst}
\bibliography{egbib.bib}

\begin{thebibliography}{10}\itemsep=-1pt

\bibitem{austin2021program}
Jacob Austin, Augustus Odena, Maxwell Nye, Maarten Bosma, Henryk Michalewski, David Dohan, Ellen Jiang, Carrie Cai, Michael Terry, Quoc Le, et~al.
\newblock Program synthesis with large language models.
\newblock {\em arXiv preprint arXiv:2108.07732}, 2021.

\bibitem{chen2021evaluating}
Mark Chen, Jerry Tworek, Heewoo Jun, Qiming Yuan, Henrique Ponde de~Oliveira Pinto, Jared Kaplan, Harri Edwards, Yuri Burda, Nicholas Joseph, Greg Brockman, et~al.
\newblock Evaluating large language models trained on code.
\newblock {\em arXiv preprint arXiv:2107.03374}, 2021.

\bibitem{deepseek-coder}
DeepSeek.
\newblock Deepseek coder: Let the code write itself, 2023.

\bibitem{dhuliawala2023chain}
Shehzaad Dhuliawala, Mojtaba Komeili, Jing Xu, Roberta Raileanu, Xian Li, Asli Celikyilmaz, and Jason Weston.
\newblock Chain-of-verification reduces hallucination in large language models.
\newblock {\em arXiv preprint arXiv:2309.11495}, 2023.

\bibitem{floridi2020gpt}
Luciano Floridi and Massimo Chiriatti.
\newblock Gpt-3: Its nature, scope, limits, and consequences.
\newblock {\em Minds and Machines}, 30:681--694, 2020.

\bibitem{hendrycks2021measuring}
Dan Hendrycks, Steven Basart, Saurav Kadavath, Mantas Mazeika, Akul Arora, Ethan Guo, Collin Burns, Samir Puranik, Horace He, Dawn Song, et~al.
\newblock Measuring coding challenge competence with apps.
\newblock {\em arXiv preprint arXiv:2105.09938}, 2021.

\bibitem{le2023codechain}
Hung Le, Hailin Chen, Amrita Saha, Akash Gokul, Doyen Sahoo, and Shafiq Joty.
\newblock Codechain: Towards modular code generation through chain of self-revisions with representative sub-modules.
\newblock {\em arXiv preprint arXiv:2310.08992}, 2023.

\bibitem{li2022competition}
Yujia Li, David Choi, Junyoung Chung, Nate Kushman, Julian Schrittwieser, R{\'e}mi Leblond, Tom Eccles, James Keeling, Felix Gimeno, Agustin Dal~Lago, et~al.
\newblock Competition-level code generation with alphacode.
\newblock {\em Science}, 378(6624):1092--1097, 2022.

\bibitem{mirzayanov2020codeforces}
Mike Mirzayanov, Oksana Pavlova, Pavel MAVRIN, Roman Melnikov, Andrew Plotnikov, Vladimir Parfenov, and Andrew Stankevich.
\newblock Codeforces as an educational platform for learning programming in digitalization.
\newblock {\em Olympiads in Informatics}, 14(133-142):14, 2020.

\bibitem{nori2023can}
Harsha Nori, Yin~Tat Lee, Sheng Zhang, Dean Carignan, Richard Edgar, Nicolo Fusi, Nicholas King, Jonathan Larson, Yuanzhi Li, Weishung Liu, et~al.
\newblock Can generalist foundation models outcompete special-purpose tuning? case study in medicine.
\newblock {\em arXiv preprint arXiv:2311.16452}, 2023.

\bibitem{team2023gemini}
Gemini Team, Rohan Anil, Sebastian Borgeaud, Yonghui Wu, Jean-Baptiste Alayrac, Jiahui Yu, Radu Soricut, Johan Schalkwyk, Andrew~M Dai, Anja Hauth, et~al.
\newblock Gemini: a family of highly capable multimodal models.
\newblock {\em arXiv preprint arXiv:2312.11805}, 2023.

\bibitem{vaswani2017attention}
Ashish Vaswani, Noam Shazeer, Niki Parmar, Jakob Uszkoreit, Llion Jones, Aidan~N Gomez, {\L}ukasz Kaiser, and Illia Polosukhin.
\newblock Attention is all you need.
\newblock {\em Advances in neural information processing systems}, 30, 2017.

\bibitem{wang2022self}
Xuezhi Wang, Jason Wei, Dale Schuurmans, Quoc Le, Ed Chi, Sharan Narang, Aakanksha Chowdhery, and Denny Zhou.
\newblock Self-consistency improves chain of thought reasoning in language models.
\newblock {\em arXiv preprint arXiv:2203.11171}, 2022.

\end{thebibliography}
}

\clearpage

\onecolumn

\appendix

\begin{appendices}

\section{Typical HumanEval code problem}
\label{app:human_eval}
\lstset{
language=C,                 
morekeywords={*,...},                       
rulecolor=\color{black},         
showspaces=false,               
showstringspaces=false,          
showtabs=false,                  
stepnumber=1,                    
stringstyle=\color{mymauve},     
title=\lstname,
        basicstyle=\ttfamily,
        showstringspaces=false,
        breaklines=true,
        keywordstyle={},
        frame=none,
        rulecolor=\color{gray},
        escapechar=|
}
\hrule
\vspace{-7mm}
\begin{lstlisting}[language=C++]
/*
Check if in given vector of numbers, are any two numbers closer to each other than given threshold. >>> 
has_close_elements({1.0, 2.0, 3.0}, 0.5) false >>> 
has_close_elements({1.0, 2.8, 3.0, 4.0, 5.0, 2.0}, 0.3) true 
*/ 
#include<stdio.h> 
#include<vector> 
#include<math.h> 
using namespace std; 
bool has_close_elements(vector<float> numbers, float threshold){

\end{lstlisting}
\hrule

\section{Why YAML output is better suited for code generation tasks than JSON output}
\label{app:yaml_vs_json}
\begin{figure*}[hbt!]
    \centering
    \begin{subfigure}[t]{0.5\textwidth}
        \centering
        \includegraphics[scale=0.35]{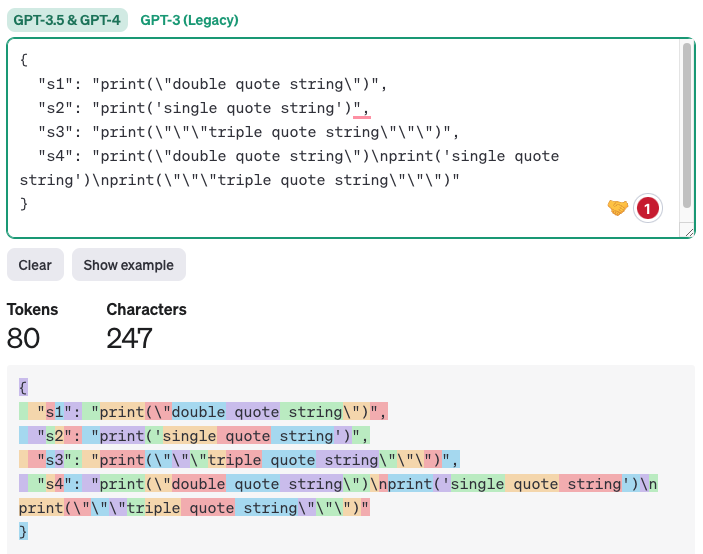}
        \caption{Token counting with JSON output}
    \end{subfigure}%
    ~ 
    \begin{subfigure}[t]{0.5\textwidth}
        \centering
        \includegraphics[scale=0.33]{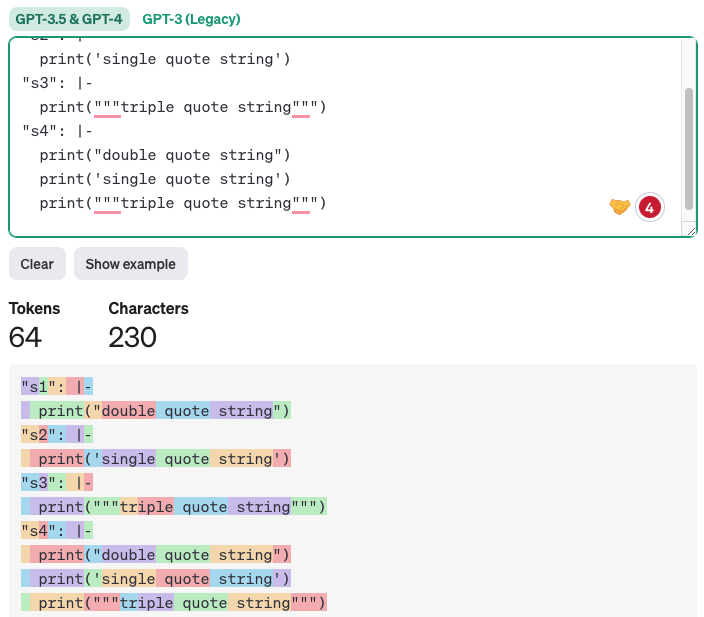}
        \caption{Token counting with YAML output}
    \end{subfigure}
    \caption{Comparison of the same output, once in JSON format, and once in YAML format. Taken from OpenAI \href{https://platform.openai.com/tokenizer}{playground}.}
\label{fig:yaml_vs_json}    
\end{figure*}

While newer GPT versions\footnote{\url{https://platform.openai.com/docs/guides/text-generation/json-mode}} have inherent support for JSON-style output, we argue that YAML output is far better for code generation. 
Why - generated code often contains single-quote, double-quote, special characters, and so on. LLMs will struggle to \textit{validly} place these characters inside a JSON format, since a JSON output needs to be surrounded with initial double quotes (see Figure~\ref{fig:yaml_vs_json} (a)). In contrast, YAML output with block scaler\footnote{\url{https://yaml-multiline.info/}} must only obey indention. Any text or code with proper indention will be a legal one (see Figure~\ref{fig:yaml_vs_json} 
 (b)).

In addition, as can be seen in Figure~\ref{fig:yaml_vs_json}, since YAML format doesn't need curly brackets, quotations or escape characters, its output has fewer tokens than JSON, hence reducing cost and inference time, and resulting in increased quality as the model needs to pay attention to fewer tokens that are not essential.

\section{What didn't work for us}
\label{app:what_didnt_work}
\begin{enumerate}
  \item 
  \textbf{Injecting the failed execution trace to the prompt:} In addition to giving the model the error message when doing iterative fixing, we also tried giving it the trace of the last X (50) lines executed. We have not observed improvement from this.
  \item
    \textbf{Injecting the last K failed code solutions to the prompt:} When doing iterative fixing, we tried injecting the last K failed code solutions to the prompt, in order to steer the model in different directions. We have not observed improvement from this.
  \item
  \textbf{Injecting the last git patch diff to the prompt:} When doing iterative fixing, we also tried to give the last applied code patch diff to the prompt. No improvement was seen.
    \item 
  \textbf{Complicated single-stage prompts:} we have not observed any significant improvement in results when trying to manipulate and optimize a single-stage prompt, or a chain of non-iterative prompts. The model still struggles to understand the lengthy problem description, tends to ignore specific details, and consistently produces wrong code.

\end{enumerate}

\section{Estimating distance between tests’ outputs}
\label{app:distance_between_tests}
When we run a solution code against an input test, the code generates an output. We compare this output to the expected output, and end up with a boolean answer: \textit{pass} or \textit{failed}. However, it is also beneficial to try and \textbf{estimate} in some sense the distance between the generated output and the correct output. 
For this matter, we employed the following logic:
\begin{itemize}
    \item 
If the test output is a number, calculate the L2 distance.
    \item 
If the test output is an array of numbers, calculate the sum of L2 distances between corresponding array cells.
    \item 
If the test output is an array of strings, compare each cell separately (boolean comparison), and return the number of non-identical cells.
\end{itemize}

This methodology enables us to produce a distance between each generated output and the correct output on CodeContests.
\end{appendices}

\end{document}